\documentclass[10pt,onecolumn,letterpaper]{article}

\usepackage{cvpr}              

\definecolor{cvprblue}{rgb}{0.21,0.49,0.74}
\usepackage[pagebackref,breaklinks,colorlinks,allcolors=cvprblue]{hyperref}

\def\paperID{*****} 
\def\confName{CVPR}
\def\confYear{2025}

\title{Open-Medical-R1: How to Choose Data for RLVR Training at Medicine Domain}

\author{
Zhongxi Qiu$^1$
\and
Zhang Zhang
\and
Yan Hu$^1$
\and
Heng Li$^1$
\and
Jiang Liu$^1$\\
$^1$Research Institute of Trustworthy Autonomous Systems\\Southern University of Science and Technology, 518055, Shenzhen, China\\
}

\begin{document}
\maketitle

\begin{abstract}
This paper explores optimal data selection strategies for Reinforcement Learning with Verified Rewards (RLVR) training in the medical domain. While RLVR has shown exceptional potential for enhancing reasoning capabilities in large language models, most prior implementations have focused on mathematics and logical puzzles, with limited exploration of domain-specific applications like medicine. We investigate four distinct data sampling strategies from MedQA-USMLE: random sampling (baseline), and filtering using Phi-4, Gemma-3-27b-it, and Gemma-3-12b-it models. Using Gemma-3-12b-it as our base model and implementing Group Relative Policy Optimization (GRPO), we evaluate performance across multiple benchmarks including MMLU, GSM8K, MMLU-Pro, and CMMLU. Our findings demonstrate that models trained on filtered data generally outperform those trained on randomly selected samples. Notably, training on self-filtered samples (using Gemma-3-12b-it for filtering) achieved superior performance in medical domains but showed reduced robustness across different benchmarks, while filtering with larger models from the same series yielded better overall robustness. These results provide valuable insights into effective data organization strategies for RLVR in specialized domains and highlight the importance of thoughtful data selection in achieving optimal performance. You can access our \href{https://github.com/Qsingle/open-medical-r1}{repository} to get the codes.
\end{abstract}    
\section{Introduction}
\label{sec:intro}
Large Language Models (LLMs) have achieved remarkable progress in recent years, with chain-of-thought techniques enabling significant performance improvements across domains including coding\cite{jain2024livecodebench,team2025kimi}, mathematics\cite{lightman2023let,team2025kimi}, and others. Reinforcement Learning with Verified Rewards (RLVR) has recently demonstrated exceptional potential for enhancing LLMs' reasoning capabilities\cite{luong2024reft}. DeepSeek-R1-Zero represents a breakthrough in this field, showing that models trained solely with RLVR can achieve excellent performance without requiring extensive reasoning data collection\cite{deepseekai2025deepseekr1incentivizingreasoningcapability}, thereby simplifying the development of chain-of-thought capabilities for downstream tasks.

Several research efforts have attempted to replicate DeepSeek-R1-Zero's success, including Huggingface's Open-R1\cite{openr1}, TinyZero\cite{tinyzero}, Logic-RL\cite{xie2025logicrlunleashingllmreasoning}, SimpleRL-Reasoning\cite{zeng2025simplerlzooinvestigatingtamingzero}, and Open-Reasoner-Zero\cite{hu2025openreasonerzeroopensourceapproach}. However, these studies predominantly focus on mathematics\cite{openr1,zeng2025simplerlzooinvestigatingtamingzero,hu2025openreasonerzeroopensourceapproach}, countdown problems\cite{tinyzero}, and logical puzzles\cite{xie2025logicrlunleashingllmreasoning}, with limited exploration of domain-specific applications such as medicine.

Some pioneering work has begun to address this gap. The Microsoft team implemented Med-RLVR using the PPO algorithm and demonstrated an 8\% accuracy improvement over supervised fine-tuning (SFT)\cite{zhang2025med}. Lai et al. introduced Group Relative Policy Optimization (GRPO)\cite{shao2024deepseekmath} for multi-modal medical diagnosis across 8 imaging modalities and 5 key diagnostic tasks\cite{lai2025med}. In a GitHub repository\cite{med-r1}, developers fine-tuned Baichuan-M1-14B\cite{baichuan-m1-2025} using GRPO on 4,000 samples with impressive results. Huang et al. achieved 6-11\% performance improvements by training on just 500 samples\cite{huang2025o1replicationjourney}.

Despite these advances in applying RLVR to medicine, a critical question remains underexplored: what is the optimal approach to organizing and selecting training data for medical RLVR? The quality, complexity, and composition of training data significantly influence a model's ability to develop robust reasoning patterns. In our previous repository \cite{Qiu_Open-Medical-R1}, we explored combining data from MedQA-USMLE\cite{jin2020disease} and MedXpertQA\cite{zuo2025medxpertqa}, discovering that complex questions facilitate chain-of-thought reasoning and help models achieve an "Aha Moment" where they self-validate answers. However, utilizing MedXpertQA, which serves as a benchmark dataset, risks knowledge leakage and potentially compromises evaluation reliability.

This observation raises fundamental questions that form the core of our current investigation: Can we sample directly from MedQA-USMLE to achieve comparable or superior results? What sampling strategies would be most effective for optimizing model performance while maintaining generalizability? How does the selection of training data impact performance across both domain-specific and general reasoning tasks?

To address these questions systematically, we selected Gemma-3-12b-it as our base model and implemented four distinct sampling strategies from MedQA-USMLE for our training dataset construction: random sampling (as a baseline approach), Phi-4\cite{abdin2024phi} filtering, Gemma-3-27b-it\cite{team2025gemma} filtering, and Gemma-3-12b-it filtering. These strategies represent different approaches to identifying appropriate training examples based on difficulty and relevance. We hypothesized that strategic sample selection could significantly enhance learning efficiency and reasoning capabilities beyond what random sampling achieves.

To comprehensively evaluate the effectiveness of our various data sampling approaches, we conducted extensive testing across multiple benchmarks including MMLU\cite{hendrycks2020measuring}, GSM8K\cite{cobbe2021gsm8k}, MMLU-Pro\cite{wang2024mmlu}, and CMMLU\cite{li2023cmmlu}. These benchmarks were specifically selected to assess both domain-specific medical reasoning capabilities and general reasoning performance, allowing us to understand the broader impacts of our sampling strategies beyond the medical domain and evaluate robustness across different types of reasoning tasks.



In conclusion, our work makes two primary contributions:
\begin{enumerate}
\item We explored and compared four distinct data sampling strategies for RLVR in the medical domain: random sampling as a baseline approach, and three model-based filtering methods using Phi-4, Gemma-3-27b-it, and Gemma-3-12b-it. 
\item We conducted comprehensive evaluations across multiple benchmarks including MMLU, GSM8K, MMLU-Pro, and CMMLU to assess both domain-specific medical reasoning capabilities and general reasoning performance. This multi-faceted evaluation approach provides a more complete understanding of how sampling strategies affect different aspects of model performance beyond domain-specific tasks.
\end{enumerate}
\section{Method}
\subsection{Group Relative Policy Optimization}
Group Relative Policy Optimization (GRPO)\cite{shao2024deepseekmath} represents an innovative reinforcement learning algorithm that utilizes relative performance metrics within action groups to optimize policy parameters. This methodology distinguishes itself from conventional policy gradient approaches such as Proximal Policy Optimization (PPO) by eliminating the requirement for a distinct value function or critic network, thus streamlining the learning process and enhancing computational efficiency.

The fundamental principle underlying GRPO is its utilization of action groups sampled from the current policy for any given state. Rather than depending on estimated value functions to determine action advantages, GRPO computes advantages relative to the mean reward within the sampled group. These relative advantages subsequently guide policy updates, reinforcing actions that exceed the group average performance while suppressing underperforming alternatives.

The GRPO algorithm can be formally articulated as follows:

Action Sampling: For a given state $s$, sample a group of $G$ actions, denoted as $a_1, a_2, ..., a_G$, from the current policy $\pi_\theta(a|s)$, where $\theta$ represents the policy parameters.
Reward Evaluation: Execute each sampled action $a_i$ in the environment and observe the corresponding reward $r_i$.
Relative Advantage Calculation: Compute the relative advantage $\hat{A}_i$ for each action $a_i$ as:
\begin{equation}
\hat{A}i = r_i - \frac{1}{G} \sum_{j=1}^{G}r_j
\end{equation}
This formulation quantifies the advantage of action $a_i$ relative to the group's mean reward. In certain implementations, this advantage may be normalized by the standard deviation of rewards within the group to improve stability.
Policy Update: Update the policy parameters $\theta$ using gradient-based optimization techniques, incorporating a clipped objective function:
\begin{align}
J_{GRPO}(\theta) &= \mathbb{E}{s, a \sim \pi\theta} \Big[ \min \big( r(s,a) \hat{A}, \nonumber\\
&\quad \text{clip}(r(s,a), 1-\epsilon, 1+\epsilon) \hat{A} \big) \Big]
\end{align} 

where $r(s,a) = \frac{\pi_\theta(a|s)}{\pi_{\theta_{old}}(a|s)}$ represents the probability ratio between current and old policies, $\theta_{old}$ denotes the policy parameters prior to updating, and $\epsilon$ is a hyperparameter that constrains the policy update magnitude.
By evaluating actions within contextual groups and updating policies based on relative performance metrics, GRPO circumvents the complexities and potential biases associated with learning separate value functions. This methodological simplification yields more efficient and robust policy learning, particularly in environments characterized by high-dimensional state or action spaces. Ongoing research continues to explore variations of GRPO to enhance its performance across diverse reinforcement learning domains.

\subsection{Data Sampling}
\begin{figure}[htbp]
    \centering
    \includegraphics[width=0.8\textwidth]{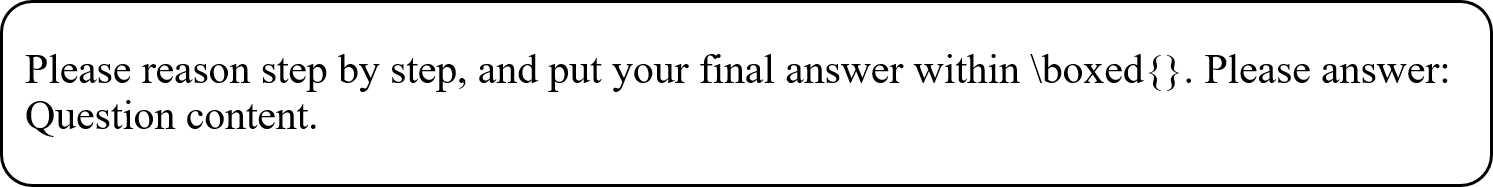}
    \caption{The prompt used to make a response for a sample.}
    \label{fig:prompt_filter}
\end{figure}

The intuitive approach would be to select samples from the dataset randomly; however, this method does not account for sample difficulty relative to model capabilities. Therefore, we developed a filtering strategy where we first had the model generate responses to candidate samples, then classified samples as "easy" if the model could answer them correctly while adhering to our prescribed format, or "hard" otherwise. Our prompt template for generating these filtering responses is shown in Fig.~\ref{fig:prompt_filter}. In our experiments, we employed Phi-4, Gemma-3-12b-it, and Gemma-3-27b-it as filtering models. To construct our training dataset, we selected 400 samples classified as hard and 100 samples classified as easy, creating a balanced yet challenging training corpus.

\subsection{Prompt}
\begin{figure}[htbp]
    \centering
    \includegraphics[width=0.8\textwidth]{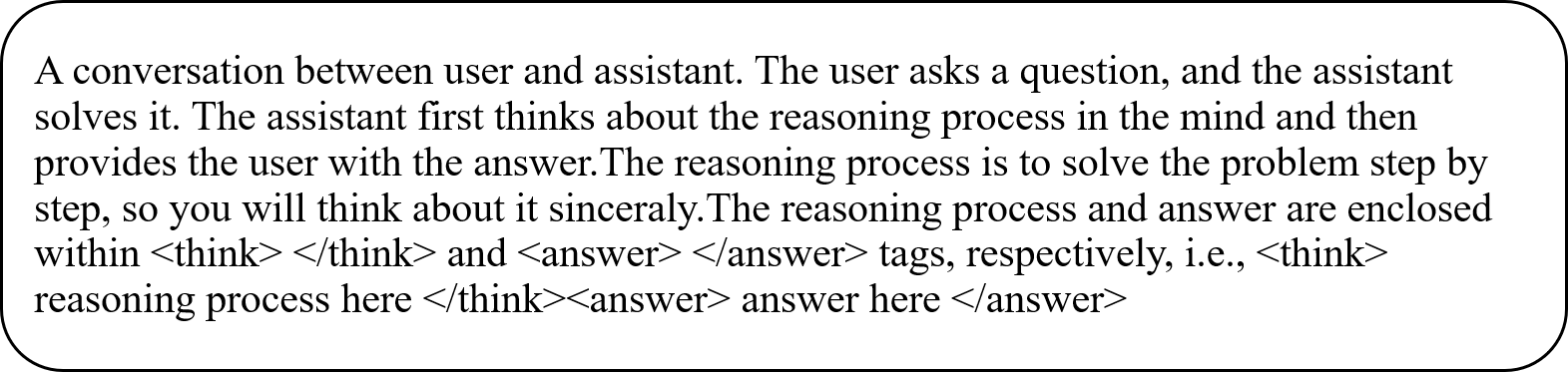}
    \caption{Template of the prompt during the training.}
    \label{fig:prompt_grpo}
\end{figure}

The prompt used to train the model is similar to the deepseek-r1-zero introduced in \cite{deepseekai2025deepseekr1incentivizingreasoningcapability}, which is shown in Fig.\ref{fig:prompt_grpo}.
\section{Experiments}
\subsection{Implementation Details}
We trained the model using the Unsloth framework \cite{unsloth} on a single NVIDIA GeForce RTX 4090 GPU with 24GB VRAM. The reward calculation incorporated three distinct components: format reward, accuracy reward, and XML count reward. We employed a learning rate of $2\times10^{-5}$ with a cosine annealing scheduler \cite{loshchilov2016sgdr} to optimize training dynamics. The implementation utilized a batch size of 3, with 3 rollouts per sample and a gradient accumulation step of 4 to effectively increase the effective batch size while managing memory constraints.

\subsection{Benchmarks}
As introduced above, we evaluated the model on MMLU, CMMLU, MMLU-Pro, and GSM8K, with results described in Table \ref{tab:benchmarks}. The results suggest that models trained on samples filtered by another model can achieve slight improvements across nearly all benchmarks compared to those trained on randomly selected samples. We do not observe the self-reflection phenomenon (Aha moment) in all of the data sampling strategies.

Interestingly, when training on data filtered by the model itself (Gemma-3-12b-it), the model achieved a relatively larger improvement on the MMLU benchmark (0.6745 vs 0.6690). However, performance on other benchmarks slightly decreased compared to the baseline. We hypothesize that while training on medical domain data improved performance in medical-related domains, the limited dataset size of only 500 samples may be insufficient, or the data selection strategy may be too simplistic.  
\begin{table*}[htbp]
    \centering
    \caption{Comparative performance metrics across MMLU, CMMLU, MMLU-Pro, and GSM8K benchmarks. The Gemma-3-12B-it model serves as our baseline reference, representing the foundation model without any GRPO-based optimization}
    \begin{tabular}{ccccc}
    \toprule
      Model   & MMLU & CMMLU & MMLU-Pro & GSM8K\\
    \midrule
       Gemma-3-12b-it & 0.6690 & 0.4750 & 0.5615 & 0.9219 \\
       GRPO(Randomly) & 0.6632 & 0.4622 & 0.5611 & 0.9151\\
       GRPO(Phi-4) & 0.6678 & 0.4699 & 0.5620 & 0.9295 \\
       GRPO(Gemma-3-27b) & 0.6699 & 0.4681 & 0.5618 & 0.9227\\
       GRPO(Gemma-3-12b) & 0.6745 & 0.4645 & 0.5577 & 0.9189\\
    \bottomrule
    \end{tabular}
    \label{tab:benchmarks}
\end{table*}

\subsection{Medical Domain Results}
\begin{figure*}[htbp]
    \centering
    \includegraphics[width=\textwidth]{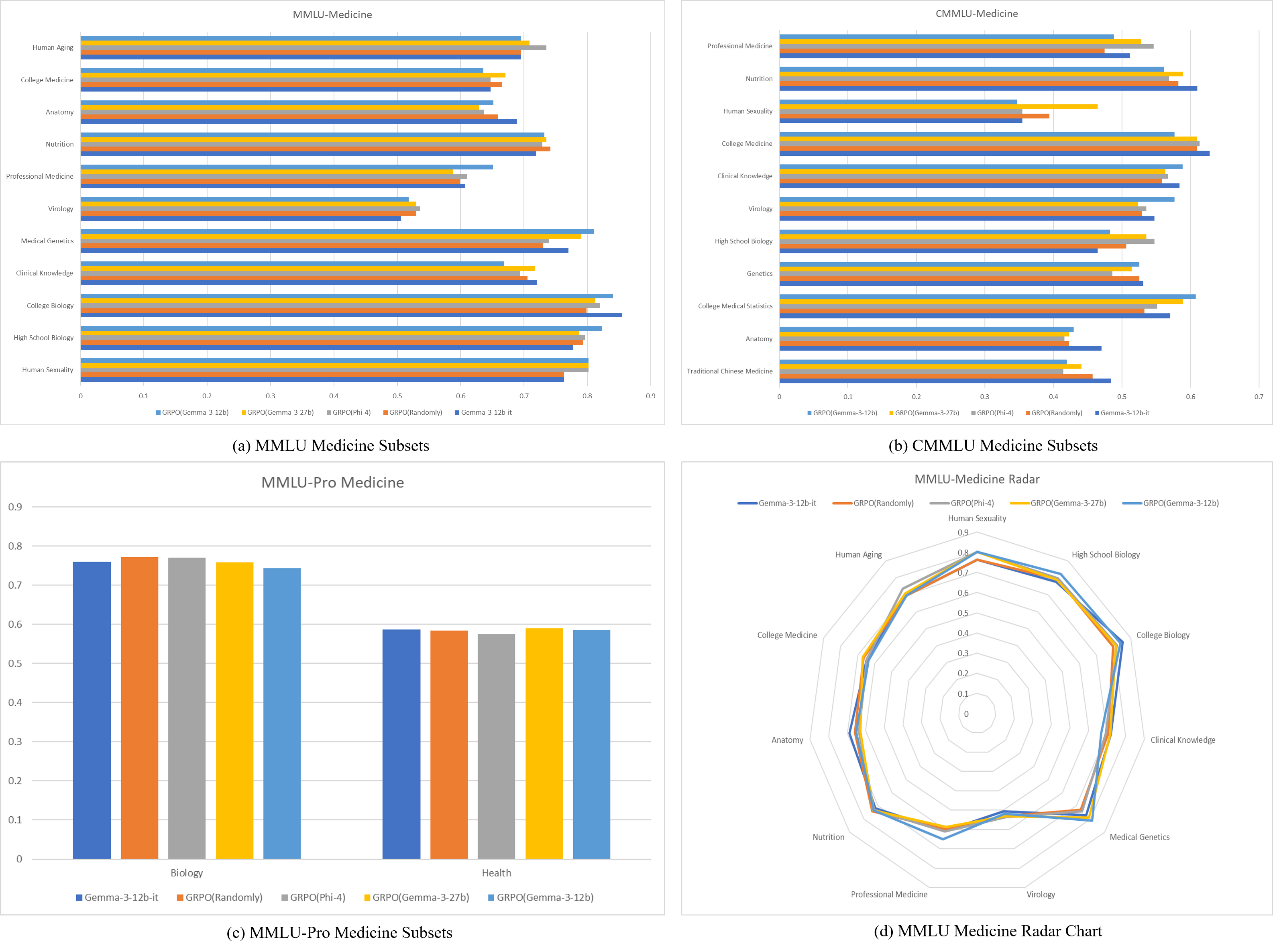}
    \caption{Results on medicine domain. (a) Results of medicine domain-related subsets from MMLU. (b) Results of medicine domain-related subsets from CMMLU. (c) Results of medicine domain-related subsets from MMLU-Pro. (d) Radar chart of the results}
    \label{fig:medicine}
\end{figure*}

Our main aim was to improve performance in medical-related domains; we illustrate the results for medicine-related fields in Fig. \ref{fig:medicine}. Similar to the benchmark results, models trained on filtered data outperformed those trained on randomly selected data. While multi-linguistic ability slightly decreased compared to the baseline, the high school biology, medical genetics, and professional medicine fields from the MMLU benchmark showed the greatest improvements when trained on data filtered by the model itself, as shown in Fig. \ref{fig:medicine}(a). However, as illustrated in Figures \ref{fig:medicine}(b) and \ref{fig:medicine}(c), models trained on self-filtered samples demonstrated insufficient robustness. In contrast, models trained on samples filtered by larger models from the same series achieved relatively better results and greater robustness across different benchmarks and languages.

\section{Conclusion}
We conducted experiments on five benchmarks, which demonstrated that models trained on samples filtered by the model itself achieved superior performance across numerous benchmarks and sub-domains compared to random selection approaches. Data filtered by the model itself facilitated optimal performance in the trained domain, though with potentially compromised robustness. Using the same architecture but with a larger model yielded better robustness. However, our current data sampling strategy remains relatively simplistic, suggesting the need for a more comprehensive strategy based on our filtering method. For future work, search tool integration or other function-calling capabilities could be incorporated during the training process to enhance the model's knowledge reasoning and response generation. Additionally, full-parameter training exploration that is based on our current strategy makes sense, which we intend to report in future work. Our experiments with LoRA adapter training revealed superior performance compared to Q-LoRA (4-bit) training. Preliminary results are available in our \href{https://huggingface.co/qiuxi337/gemma-3-12b-it-grpo}{repository}.
{
    \small
    \bibliographystyle{ieeenat_fullname}
    \bibliography{main}
}

\clearpage

\section{Domain results}
\label{sec:domain results}
\subsection{MMLU}
\begin{figure}[htbp]
    \centering
    \includegraphics[width=\textwidth]{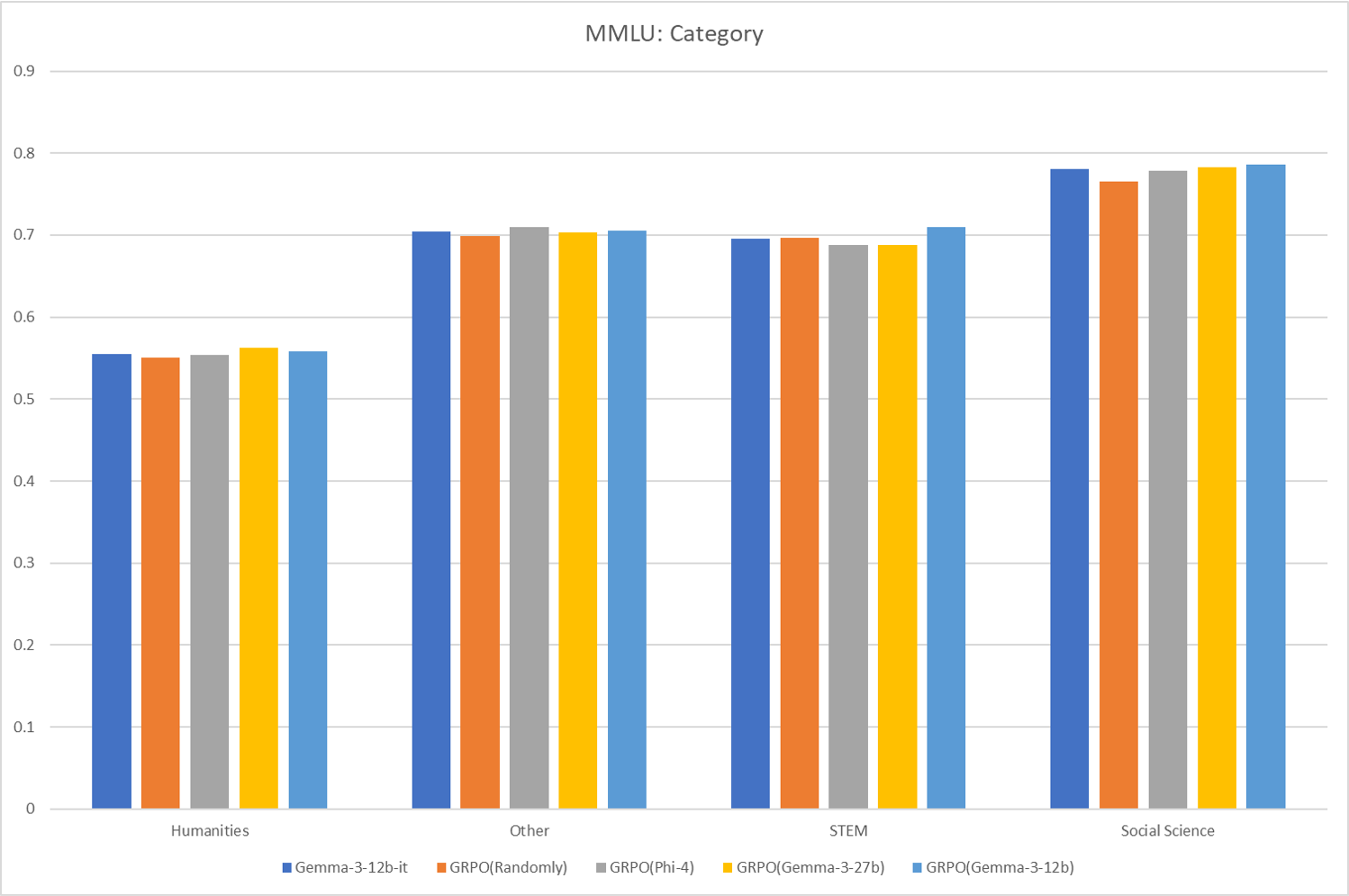}
    \caption{Results across MMLU's four main categories.}
    \label{fig:mmlu_category}
\end{figure}

\begin{figure}[htbp]
    \centering
    \includegraphics[width=\textwidth]{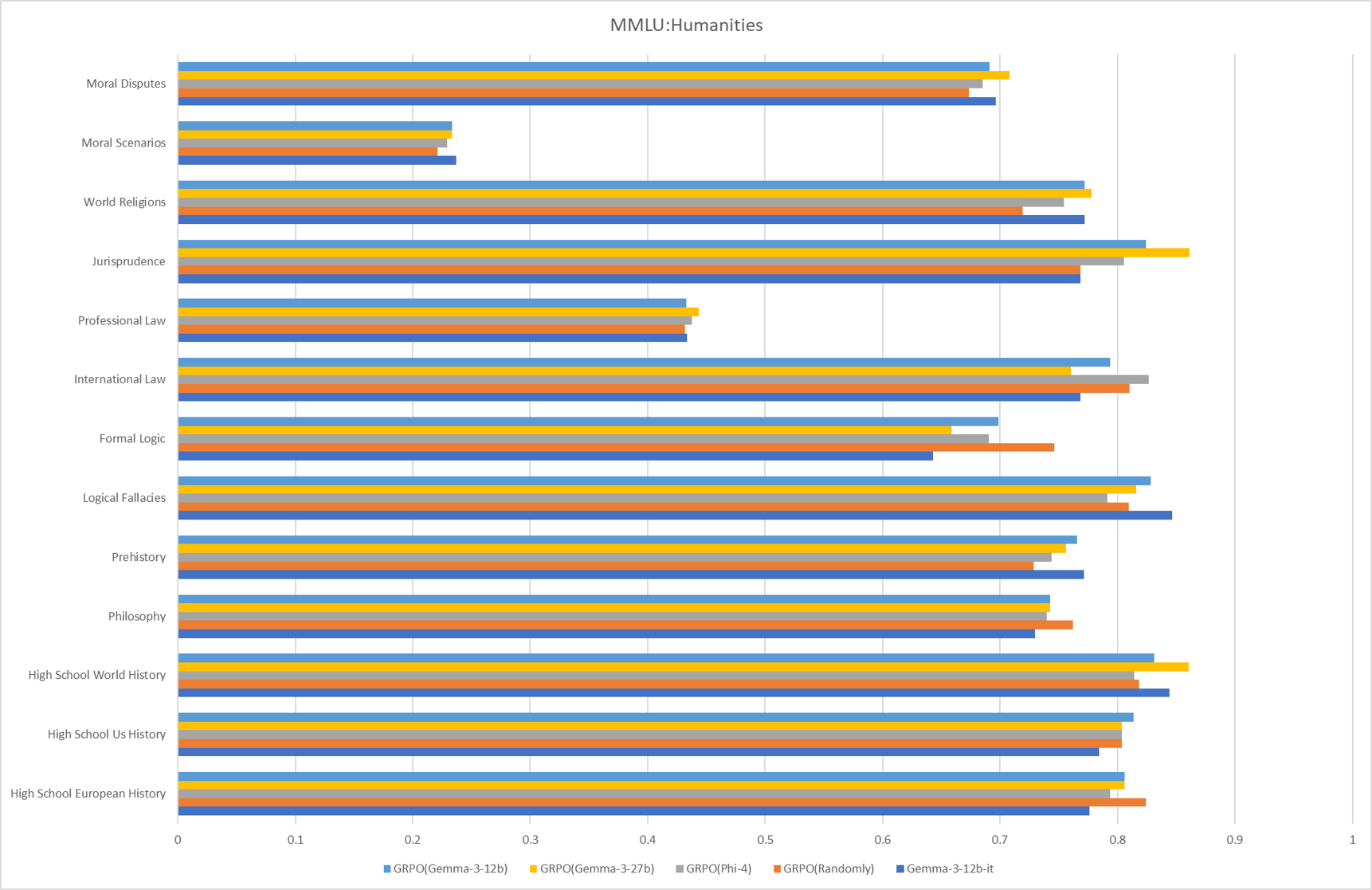}
    \caption{Performance breakdown for the Humanities category from the MMLU benchmark.}
    \label{fig:mmlu_humanities}
\end{figure}

\begin{figure}[htbp]
    \centering
    \includegraphics[width=\textwidth]{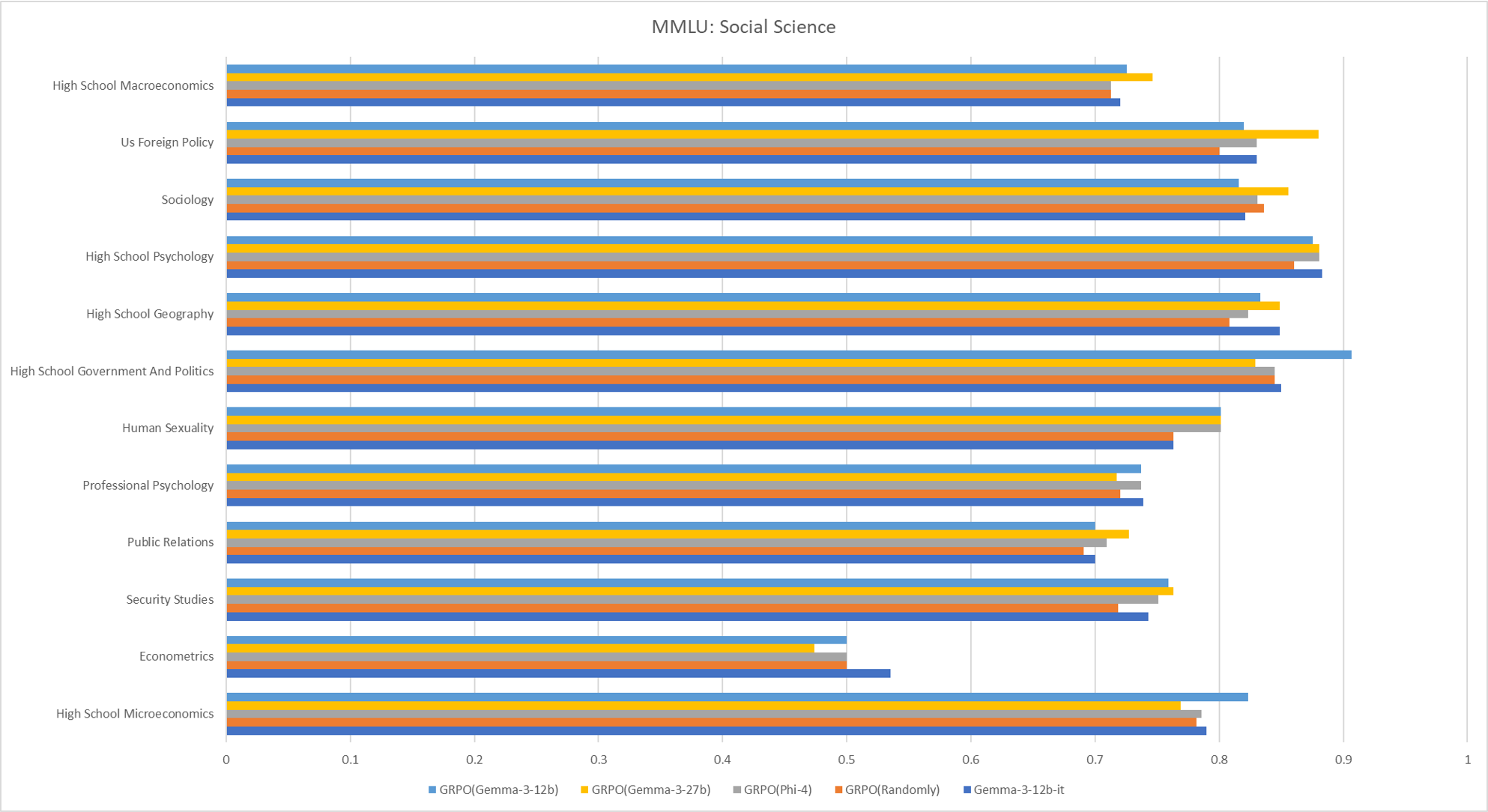}
    \caption{Performance breakdown for the Social Science category from the MMLU benchmark.}
    \label{fig:mmlu_social_science}
\end{figure}

\begin{figure}[htbp]
    \centering
    \includegraphics[width=\textwidth]{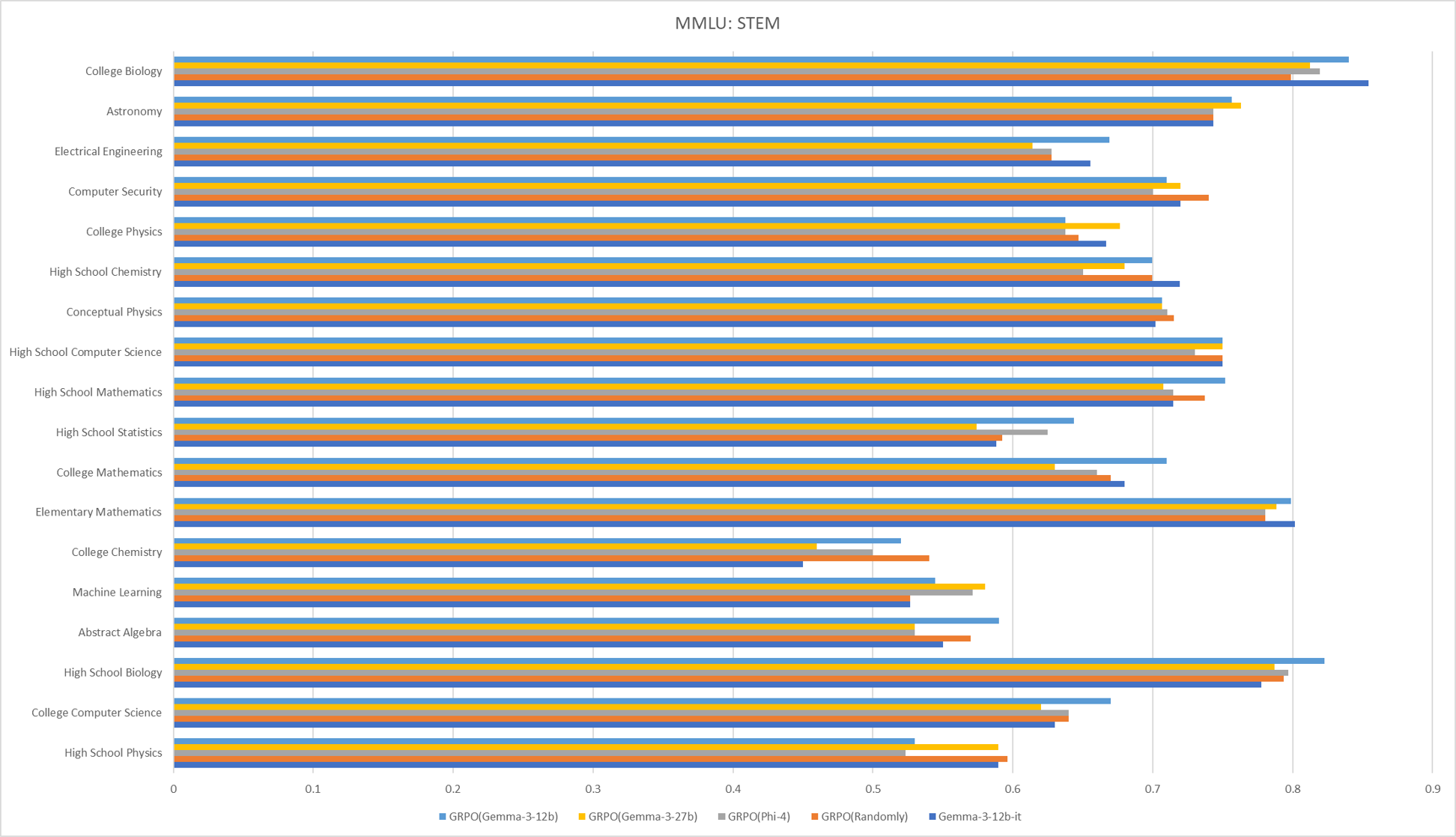}
    \caption{Performance breakdown for the STEM category from the MMLU benchmark.}
    \label{fig:mmlu_stem}
\end{figure}

\begin{figure}[htbp]
    \centering
    \includegraphics[width=\textwidth]{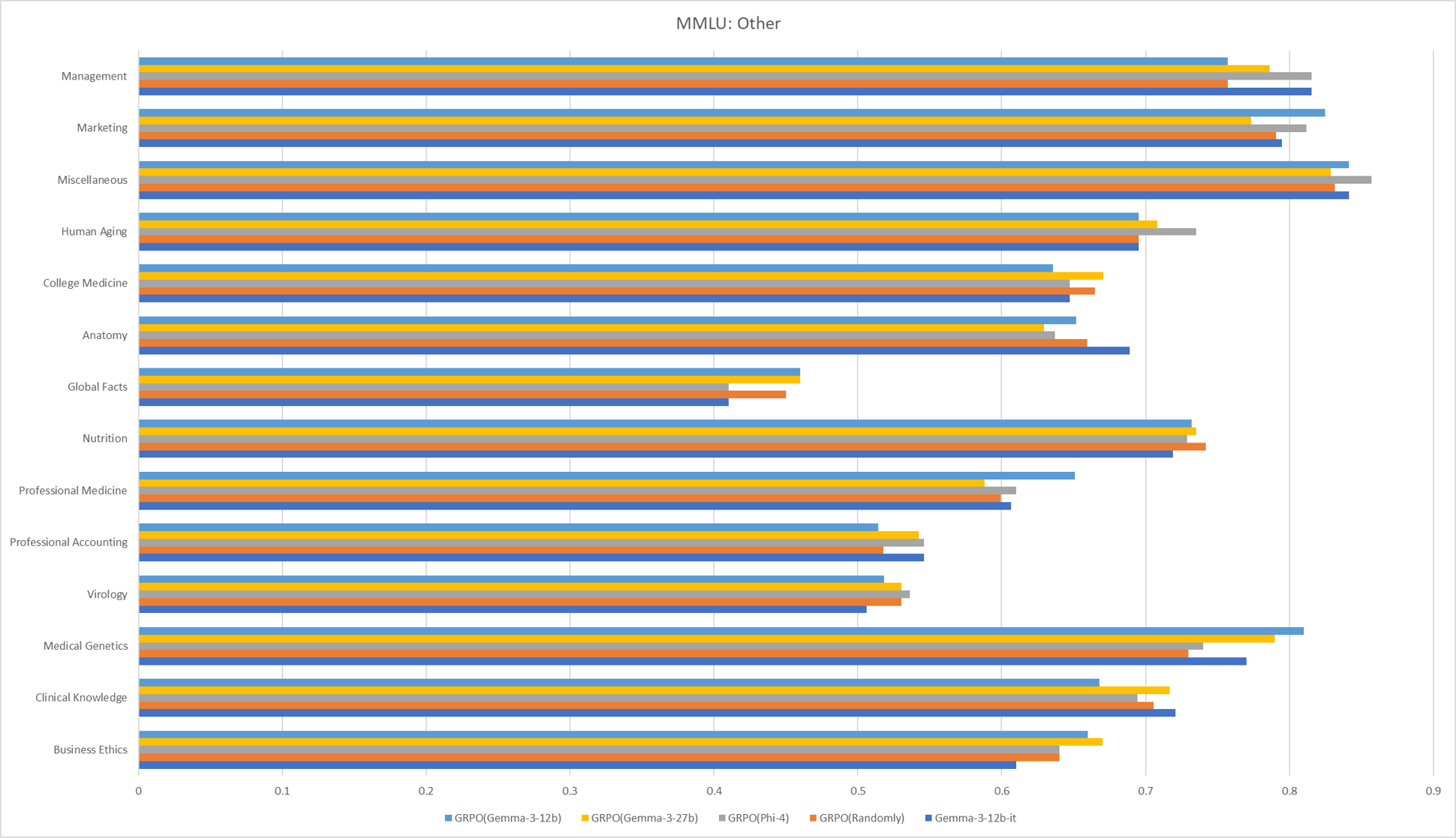}
    \caption{Performance breakdown for the Other category from the MMLU benchmark.}
    \label{fig:mmlu_other}
\end{figure}

The MMLU benchmark comprises four categories: Humanities, Social Science, STEM (Science, Technology, Engineering, Mathematics), and Other. Results across these categories are illustrated in Figure \ref{fig:mmlu_category}. Notably, the model trained on data filtered by the model itself demonstrates superior performance in the STEM category compared to alternative sampling strategies. The performance in the Humanities category is presented in Figure \ref{fig:mmlu_humanities}, while results for the Social Science category are shown in Figure \ref{fig:mmlu_social_science}. Similarly, Figures \ref{fig:mmlu_stem} and \ref{fig:mmlu_other} depict the results for the STEM and Other categories, respectively. A detailed analysis of these results reveals that GRPO training significantly enhances performance in reasoning-oriented domains, particularly mathematics.

\subsection{CMMLU}
\begin{figure}[htbp]
    \centering
    \includegraphics[width=\textwidth]{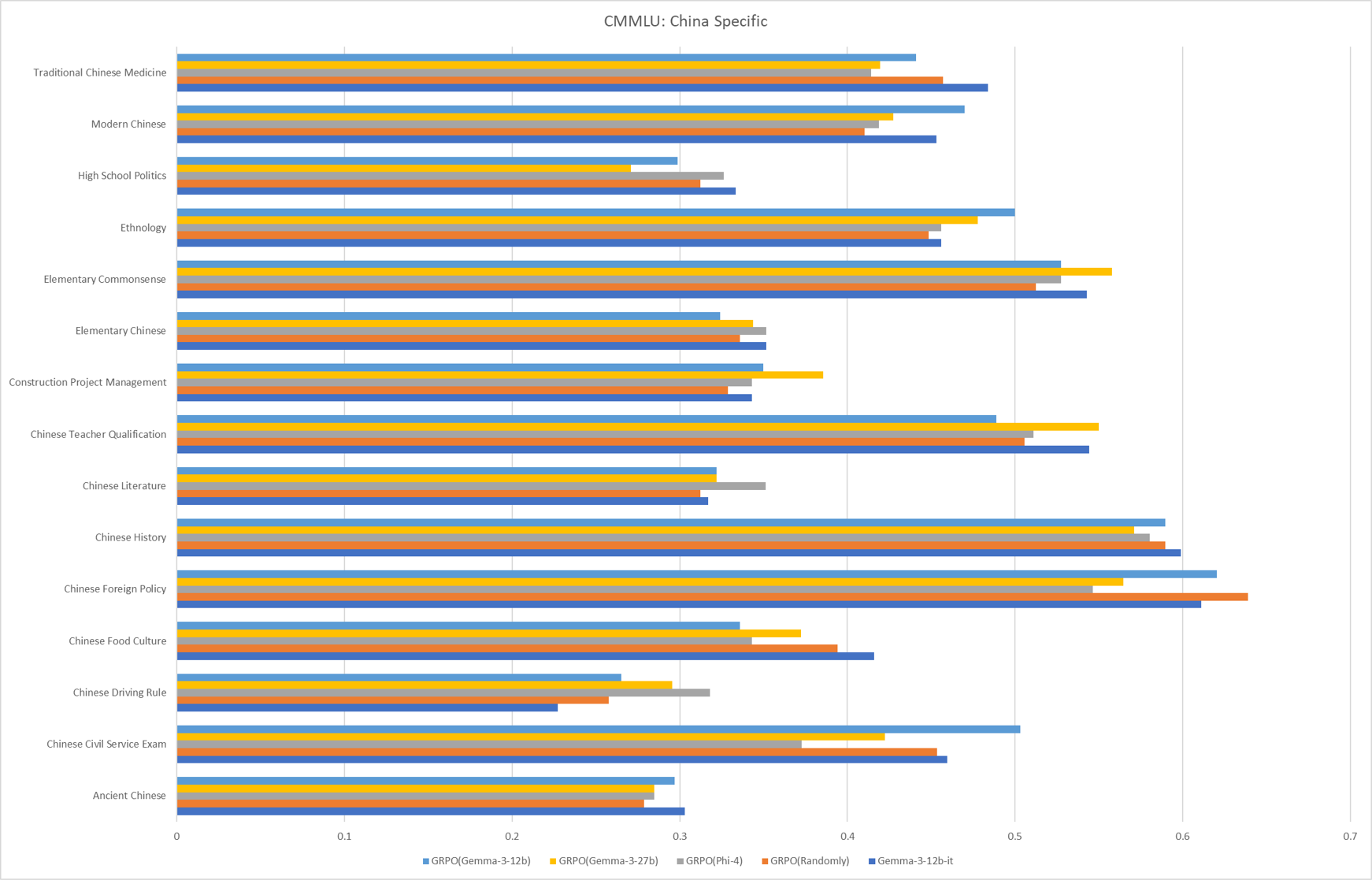}
    \caption{Performance breakdown for the China Specific category from the CMMLU benchmark.}
    \label{fig:cmmlu_china_specific}
\end{figure}

\begin{figure}[htbp]
    \centering
    \includegraphics[width=\textwidth]{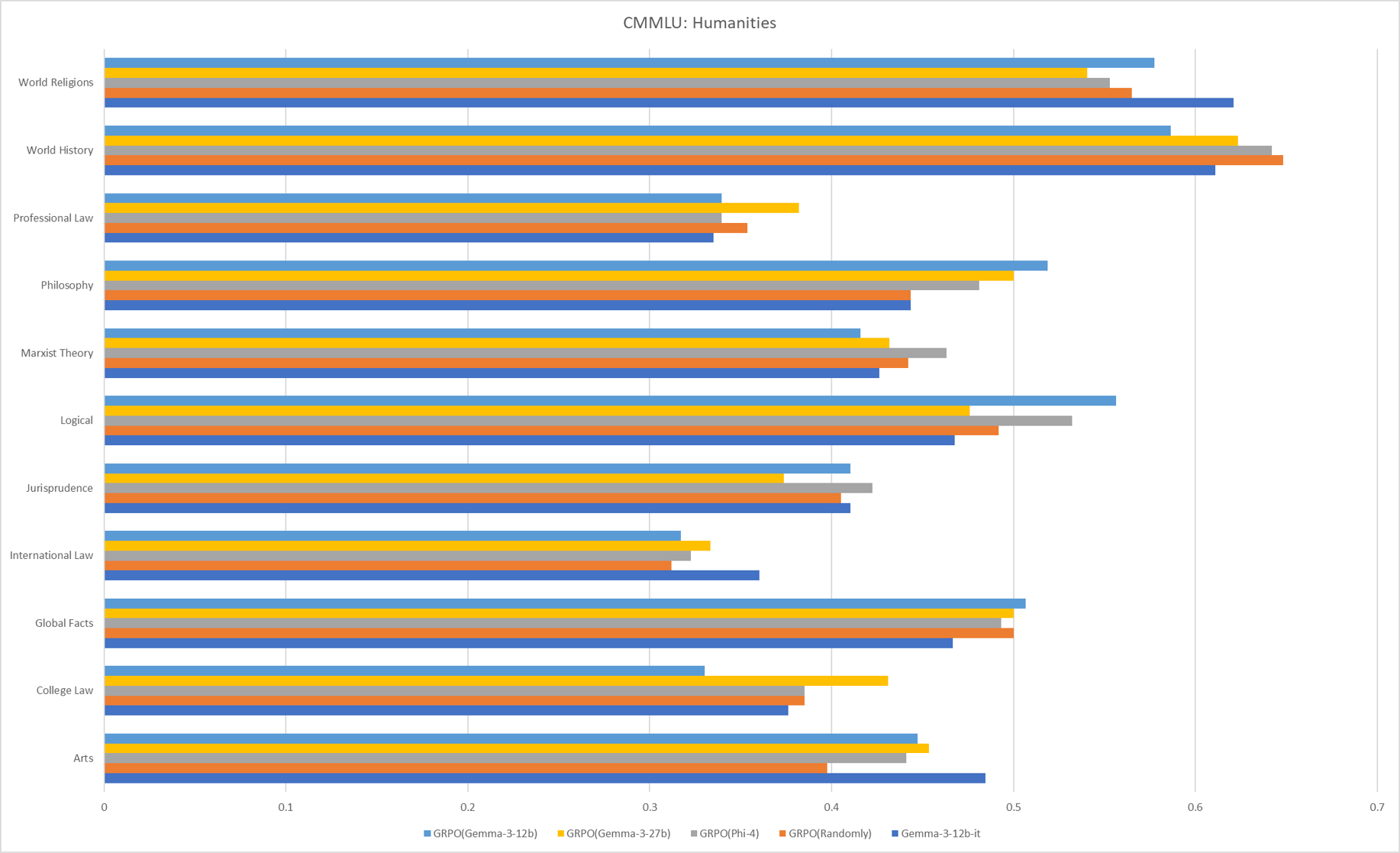}
    \caption{Performance breakdown for the Humanities category from the CMMLU benchmark.}
    \label{fig:cmmlu_humanities}
\end{figure}

\begin{figure}[htbp]
    \centering
    \includegraphics[width=\textwidth]{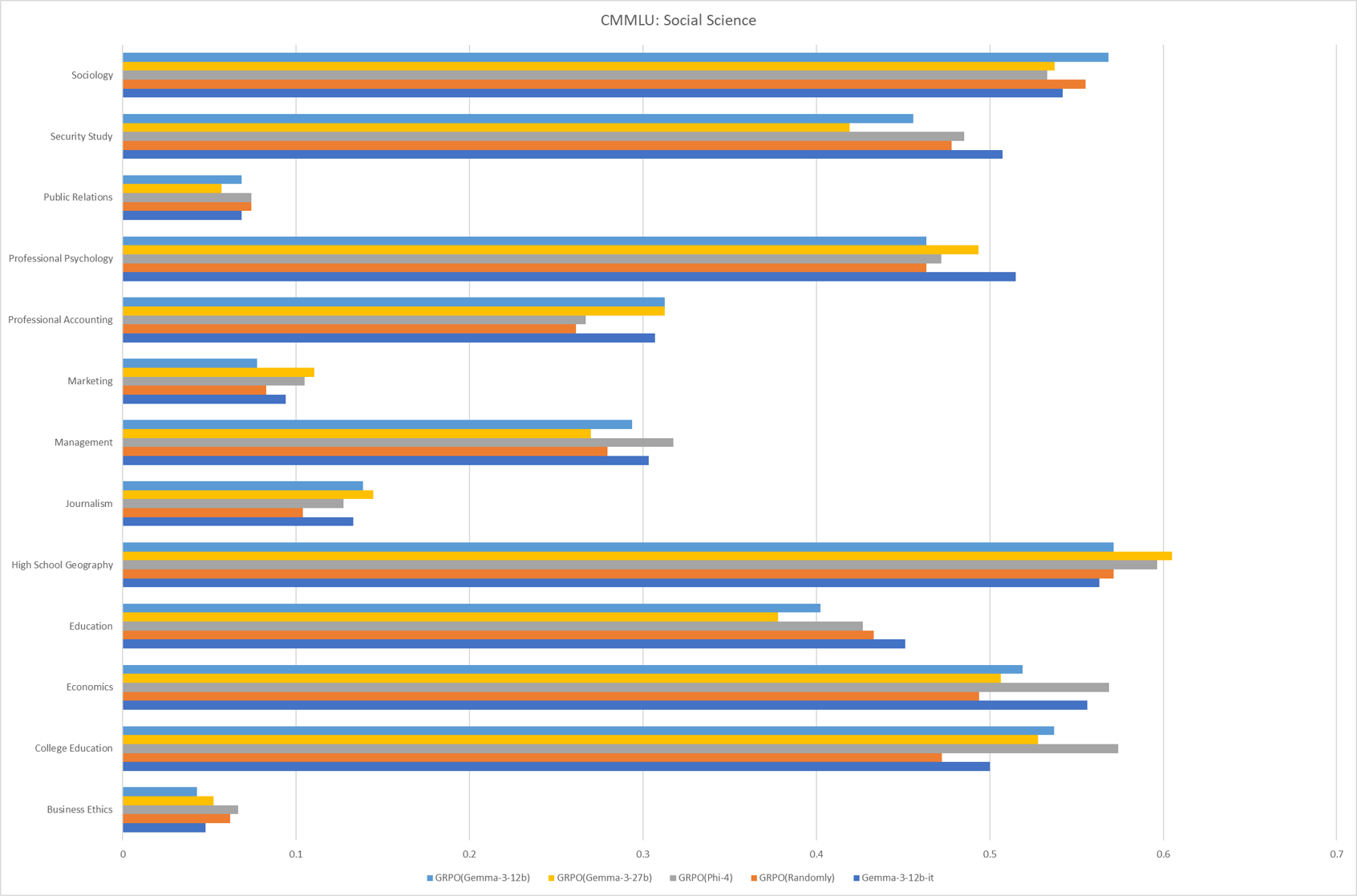}
    \caption{Performance breakdown for the Social Science category from the CMMLU benchmark.}
    \label{fig:cmmlu_social_science}
\end{figure}

\begin{figure}[htbp]
    \centering
    \includegraphics[width=\textwidth]{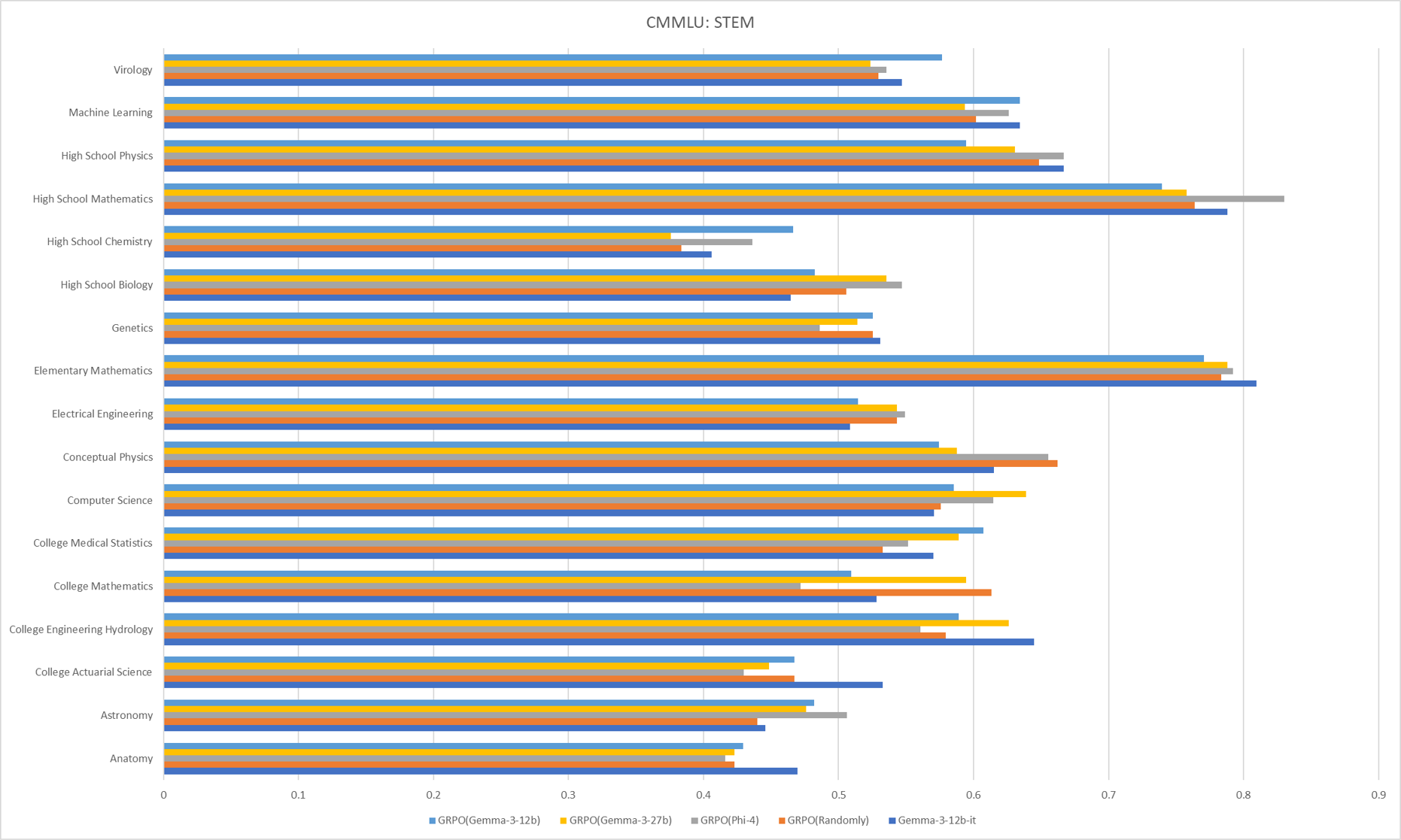}
    \caption{Performance breakdown for the STEM category from the CMMLU benchmark.}
    \label{fig:cmmlu_stem}
\end{figure}

\begin{figure}[htbp]
    \centering
    \includegraphics[width=\textwidth]{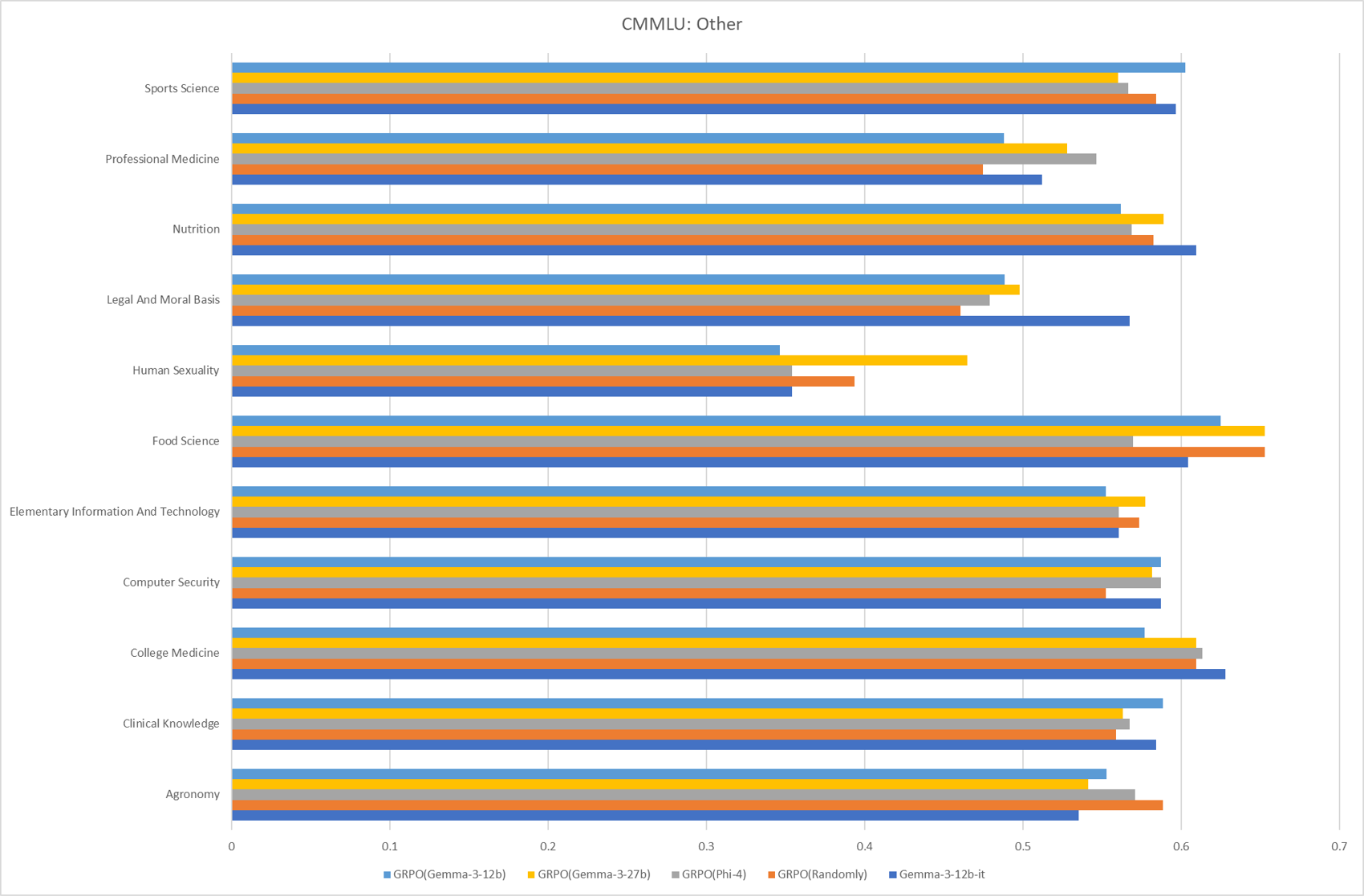}
    \caption{Performance breakdown for the Other category from the CMMLU benchmark.}
    \label{fig:cmmlu_other}
\end{figure}

The CMMLU benchmark comprises fives categories: China Specific, Humanities, Social Science, STEM (Science, Technology, Engineering, Mathematics), and Other. Results across these categories are illustrated in Figure \ref{fig:mmlu_category}. Figure \ref{fig:cmmlu_china_specific} shows the performance on China Specific category. The performance in the Humanities category is presented in Figure \ref{fig:cmmlu_humanities}, while results for the Social Science category are shown in Figure \ref{fig:cmmlu_social_science}. Similarly, Figures \ref{fig:cmmlu_stem} and \ref{fig:cmmlu_other} depict the results for the STEM and Other categories, respectively. The GRPO training on a single language data may harms the multi linguist performance, which is make sense. But trained on the data that is filtered by the Gemma-3-27b-it or Phi-4 may keep relative robustness, the performance on high-school biology, Chinese Civil Service Exam achieve a significant improvement.

\begin{figure}[htbp]
    \centering
    \includegraphics[width=\textwidth]{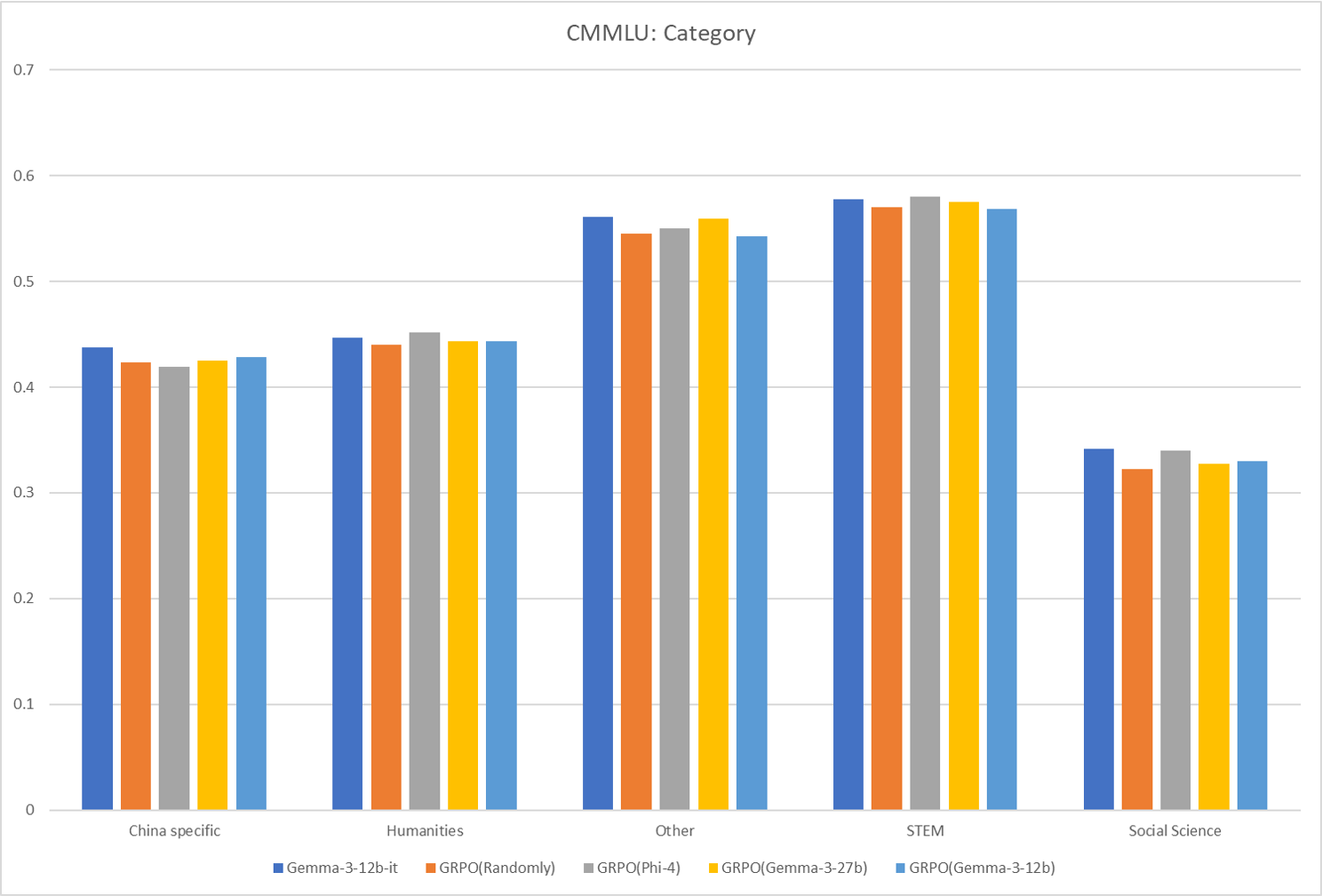}
    \caption{Results across CMMLU's five main categories.}
    \label{fig:cmmlu_category}
\end{figure}

\subsection{MMLU-Pro}
\begin{figure}[htbp]
    \centering
    \includegraphics[width=\textwidth]{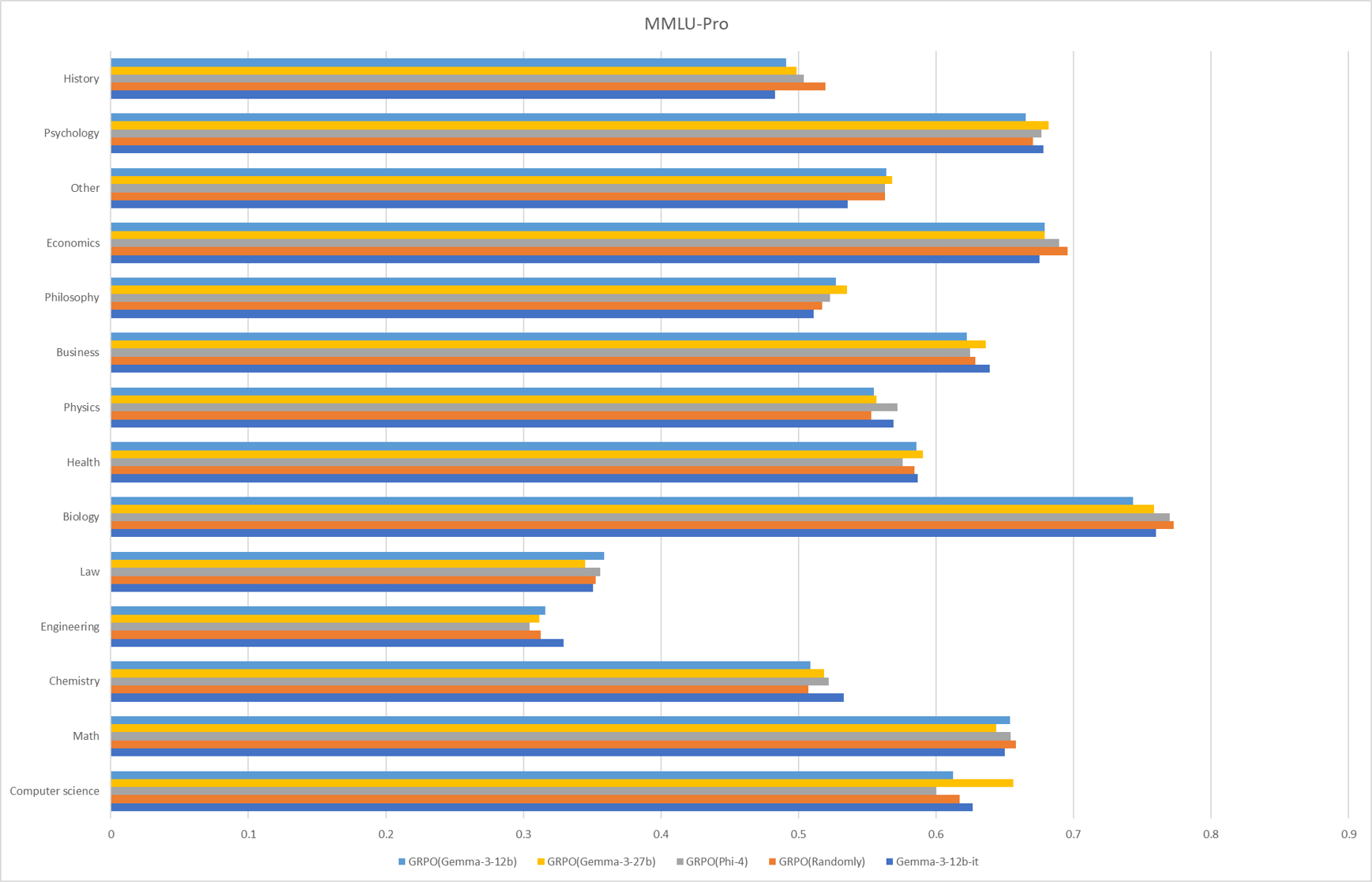}
    \caption{Detail results for MMLU-Pro benchmarks.}
    \label{fig:mmlu_pro}
\end{figure}

The detail results for the MMLU-Pro benchmark are shown in Fig.\ref{fig:mmlu_pro}

\end{document}